\documentclass[letterpaper, 10 pt, conference]{ieeeconf}  

\IEEEoverridecommandlockouts                              

\usepackage{cite}
\usepackage{amsmath,amssymb,amsfonts}
\usepackage[ruled,linesnumbered]{algorithm2e}
\usepackage{algorithmic}
\usepackage{graphicx}
\usepackage{subfigure}
\usepackage{graphics}
\usepackage{textcomp}
\usepackage{xcolor}
\usepackage{mathptmx} 
\usepackage{times} 
\usepackage{bm}
\usepackage{multirow}
\usepackage{booktabs}
\usepackage{array}
\usepackage{tabularx,booktabs}
\usepackage{overpic}
\usepackage{url}
\usepackage[bookmarks=false]{hyperref}
\usepackage{rotating}
\usepackage{mathtools}
\usepackage{multirow}

\usepackage{verbatim}

\title{\LARGE \bf Learning to Socially Navigate in Pedestrian-rich Environments\\ with Interaction Capacity
}

\author{Quecheng Qiu$^1$, Shunyi Yao$^2$, Jing Wang$^1$, Jun Ma$^1$, Guangda Chen$^2$, and Jianmin Ji$^{2, *}$ 
\thanks{
The work is partially supported by the National Key Research and Development Program of China (No. 2018AAA0100500), CAAI-Huawei MindSpore Open Fund, Anhui Provincial Development and Reform Commission 2020 and 2021 Projects, Key-Area Research and Development Program of Guangdong Province 2020B0909050001, Shenzhen Yijiahe Technology R\&D Co., Ltd., and Huawei Cloud Computing Technologies Co., Ltd.
}
\thanks{* The corresponding author.}
\thanks{$^1$ School of Data Science, University of Science and Technology of China, Hefei, 230026, China
{\tt\small \{qiuqc, wj980828, markjun\}@mail.ustc.edu.cn}.}%
\thanks{$^2$ School of Computer Science and Technology, USTC, Hefei, 230026, China
{\tt\small \{ustcysy, cgdsss\}@mail.ustc.edu.cn, jianmin@ustc.edu.cn}.}%
}%

\begin{document}
\maketitle
\thispagestyle{empty}
\pagestyle{empty}

\begin{abstract}
        Existing navigation policies for autonomous robots tend to focus on collision avoidance while ignoring human-robot interactions in social life. 
        For instance, robots can pass along the corridor safer and easier if pedestrians notice them. 
        Sounds have been considered as an efficient way to attract the attention of pedestrians, which can alleviate the freezing robot problem.
        In this work, we present a new deep reinforcement learning (DRL) based social navigation approach for autonomous robots to move in pedestrian-rich environments with interaction capacity. 
        Most existing DRL based methods intend to train a general policy that outputs both navigation actions, i.e., expected robot's linear and angular velocities, and interaction actions, i.e., the beep action, in the context of reinforcement learning. 
        Different from these methods, we intend to train the policy via both supervised learning and reinforcement learning. 
        In specific, we first train an interaction policy in the context of supervised learning, which provides a better understanding of the social situation, then we use this interaction policy to train the navigation policy via multiple reinforcement learning algorithms. 
        We evaluate our approach in various simulation environments and compare it to other methods. 
        The experimental results show that our approach outperforms others in terms of the success rate. 
        We also deploy the trained policy on a real-world robot, which shows a nice performance in crowded environments.
\end{abstract}

\section{Introduction}

Mobile robots have been widely used in various real-world applications, where the capability of autonomous navigation is critical.
Most of these applications require the robot to navigate in pedestrian-rich environments, like malls, hotels, and hospitals.
It is challenging for the robot to shuttle freely in the crowd while following social conventions.

Most existing work for robot navigation focus on collision avoidance while ignoring human-robot interactions in social life.
Some of them navigate the robot by first analyzing pedestrians' behaviors and predicting their trajectories~\cite{RVO, ORCA, SFM, phillips2011sipp}.
However, it is hard to precisely predicate pedestrians' trajectories and is like to suffer from the freezing robot problem in dense crowds.
On the other hand, deep reinforcement learning (DRL) based methods have been applied to collision avoidance with promising results~\cite{CADRL,SACADRL,liu2020robot,GCN,chen2020distributed}, where the reward functions need to be specified to describe how the robot ought to move. 
Most of them ignore human-robot interactions, regard pedestrians as moving obstacles, and plan the avoiding-path over the range of pedestrians' moving area. 
However, it would be much more efficient and safer if the robot were noticed by pedestrians, which brings new challenges for the robot on when to perform interaction actions and how to navigate in this human-aware fashion. 

Interaction via sounds has been considered as an efficient way to attract the attention of pedestrians. 
Nishimura et~al.~\cite{L2B} design a DRL framework to learn general policies that output both navigation actions, i.e., expected robot's linear and angular velocities, and interaction actions, i.e., the beep action.
However, there are several drawbacks for training these general policies in the context of DRL:
1) It is difficult to design a specific reward function that encourages the robot to perform interaction actions properly in various social environments, like a narrow corridor with moving pedestrians or a hall full of standing people. 
2) Moreover, the social conventions followed by people are hard to be specified by reward functions. 
3) The general policy for both navigation actions and interaction actions results in a much larger policy space to be explored, and the nature of the sparse reward for navigation requires a long-term memory to store the succeeding states and actions after an interaction action during the training.

In this paper, we address these problems by training the policy via both DRL and supervised learning (SL), as shown in Fig.~\ref{network}.
In specific, we first train an interaction policy in the context of SL. 
Note that, it is easier for humans to label the situations when an interaction action should be performed following social conventions, other than specifying such a reward function.
Then a proper interaction policy that is suitable for the certain social environment can be learned via SL from the labeled interaction data. 
Later, we use this learned interaction policy to train the navigation policy via DRL. 
Multiple DRL methods have been applied for robot navigation with promising results. 
We can slightly adjust these methods to train the navigation policy given the learned interaction policy. 
We implement our approach based on multiple DRL algorithms, evaluate the approach in various simulation environments, and compare it to other methods. 
The experimental results show that our approach outperforms others in terms of the success rate. 
We also deploy the trained policy on a real-world robot, which shows a nice performance in crowded environments.

Our main contributions are summarized as follows:
\begin{itemize}
	\item We propose a learning approach for social navigation with interaction capacity in pedestrian-rich environments. 
         We first train a Supervised Learning Interaction (SLI) policy, then we use this learned interaction policy to train the navigation policy via DRL.
	\item We implement our approach based on multiple DRL algorithms, which implies that the approach can be used to incorporate other DRL based navigation methods with interaction policies. 
	\item We evaluate the approach in various simulation environments and compare it to other methods. The experimental results show that our approach outperforms others in terms of the success rate. We also deploy the trained policy on a real-world robot, which shows a nice performance in crowded environments. 
\end{itemize}

\section{Related Work}

\subsection{Crowd-avoidance Robot Navigation.}

It is challenging for robots to navigate in pedestrian-rich environments since the prediction trajectories of pedestrians are hardly to be precise. 
Prior work address the problem by introducing various reciprocal methods, like SFM~\cite{SFM}, RVO~\cite{RVO}, ERVO~\cite{ERVO} and ORCA~\cite{ORCA}, which have achieved good results both in both simulation and real-world testing environments.
However, these methods rely on certain assumptions about patterns of human behavior, which can hardly handle the scenarios with dynamic groups that exhibit different spatial behavior in pedestrian-rich environments.
Other methods, like Fuzzy control~\cite{FuzzySet}, are used to predict the movement of nearby human~\cite{FuzzyControl}.

On the other hand, DRL based methods have been applied to crowd avoidance navigation tasks with promising results. 
These DRL methods can generally be divided into two categories, i.e., value-based methods and policy-based methods.
Value-based methods learn policy based on value function while policy-based methods based on gradient boost.
For both of them, accurately estimating the state-value function is vital for the navigation performance, which is highly relied on the presentation of agents' observation. 
For value-based methods, Chen et~al.~\cite{SACADRL} propose an agent-level method called SA-CADRL and Chen et al.~\cite{GCN} employ Graph Convolutional Networks for a better representation of the crowd as well.
Both of them have made promising achievements in social navigation tasks.
However, they are lack of support for continuous actions and are not good at searching for stochastic optimal policies~\cite{sutton2018reinforcement}.
As for policy-based methods, Fan et al.~\cite{fan2020distributed} propose a hybrid-RL policy for multi-agent collision avoidance without pedestrian information, which also works in dense crowds.
Liu et al.~\cite{liu2020robot} separate pedestrians from static obstacles based on the angular map.
Yao et al.~\cite{yao2021crowd} use both the egocentric sensor map and the pedestrian map as the real-time observation, which performs well in pedestrian-rich environments without human-robot interactions.

Imitation learning methods~\cite{liu2018map} aim to learn a policy as close to the expert policy as better, which requires the collection of experts' demonstrations.
Tai et~al.~\cite{GAILnav} applies GAIL~\cite{GAIL} to navigation tasks with raw depth inputs, which combines the idea of GAN~\cite{GAN} and imitation learning and performs well in realistic scenarios.

\subsection{Human-Robot Interaction in Robot Navigation.}

More and more robots are running in dynamic environments with humans, it's important for autonomous robots to interact with humans following the social conventions.
Kruse et~al.~\cite{2013survey} summarize that, comfort, naturalness, and sociability are three key requirements of social navigation.

Prior work usually use model-based methods to simulate the interaction between robots and crowds. 
Some early work regard the crowd as static obstacles~\cite{borenstein1989real,borenstein1990real,borenstein1991vector}, and avoid collision by establishing a virtual force field.
Later, the Social Force model~\cite{ferrer2013robot,ferrer2014behavior,mehta2016autonomous} is proposed.
In this model, social interactions are simulated by controlling the repulsive forces of different types (human-to-human, human-to-robot), where the force parameters usually need to be adjusted separately, which limits its generalization to many applications.
ORCA~\cite{ORCA} propose a method for reciprocal $n$-body collision avoidance under the assumption that the protocol used by each robot is the same.
Furthermore, ERVO~\cite{ERVO} proposes a new emotional contagion model on the basis of RVO~\cite{RVO}.

In contrast, rather than formulating strategies, learning-based methods have received widespread attention to guide the robot to navigate by simulating crowd behavior.
Kim and Pineau~\cite{kim2016socially} use inverse reinforcement learning to learn a cost function through human demonstrations.
In order to deal with various neighbors, Everett et al.~\cite{everett2018motion} use the LSTM model and process each neighbor in the reverse order of the distances from the robot.
However, this model cannot fully represent all the interactions between the crowd and the robot. 
SARL~\cite{SARL} designs a social attentive pooling module to encode crowd cooperative behaviors, where the human-robot and human-human interaction are jointly modeled, which makes up for the deficiencies in~\cite{everett2018motion}.
However, all of the above methods may fall into stuck or time out when the crowd is dense. 
To address the freezing robot problem, Dugas et~al.~\cite{IAN} propose a multi-model socially navigation behaviors which allows robots to speak and nudge.
Nishimura et~al.~\cite{L2B} design the safety-efficiency trade-off reward based on Sequential Social Dilemmas (SSD)~\cite{SSD}.
Both of them propose that instead of getting into the freezing robot problem, it is better to encourage robots to interact with pedestrians appropriately. 

Following above ideas, we address the social navigation problem with interaction capacity by training the policy via both DRL and SL.

\section{Approach}

It is neither safe nor efficient for a robot to take the initiative to avoid collisions with every pedestrian in a pedestrian-rich environment. 
On the other hand, with interaction capacity, the robot can interact with humans when it were going to be stuck, which can alleviate the freezing robot problem and improve efficiency.

\begin{figure}[hbt]
        \centering
        \begin{overpic}[width=\linewidth]{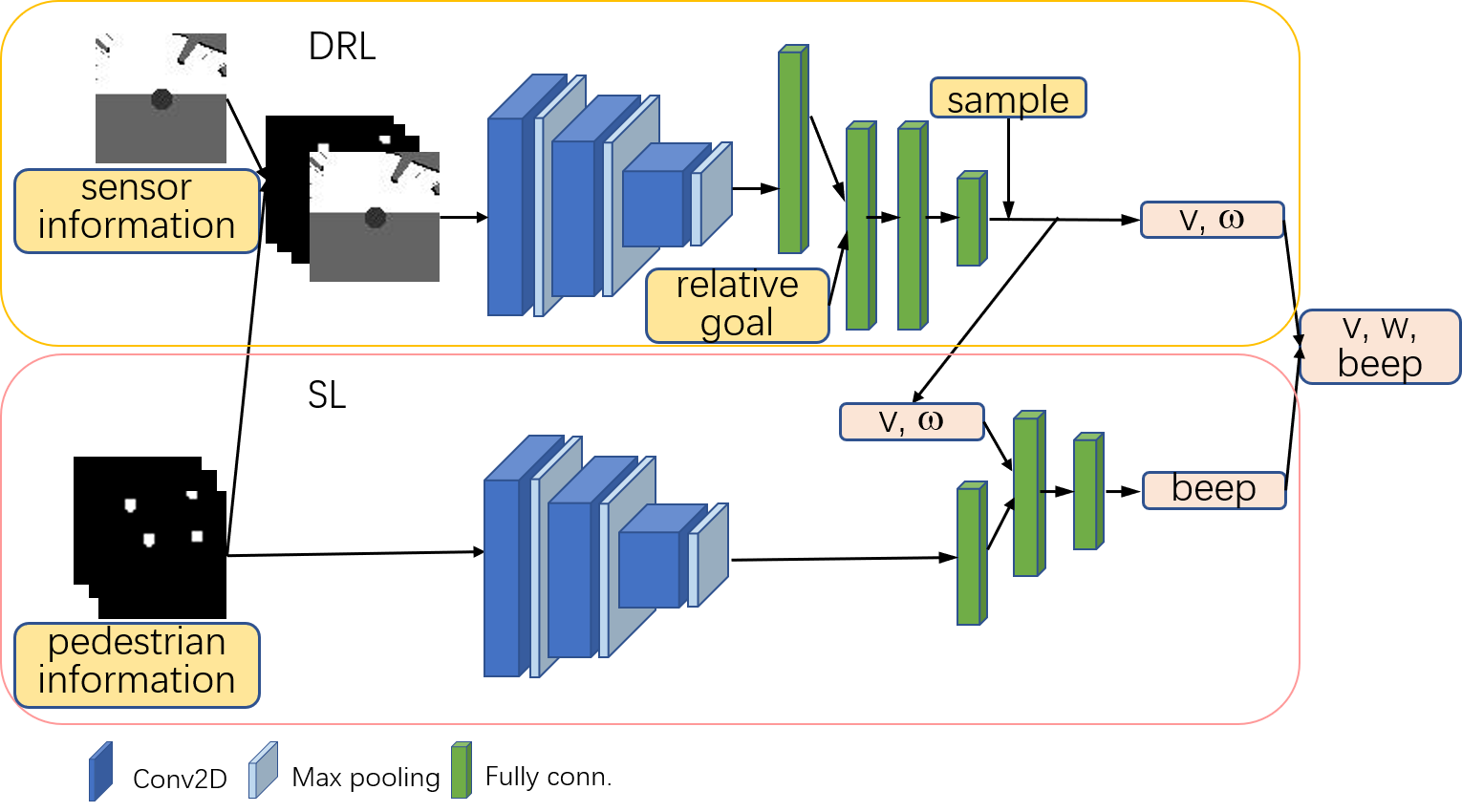}
        \end{overpic}
        \caption{The architecture of the crowd navigation policy network.}
        \label{network}
\end{figure}

In this section, we propose a new learning approach for social navigation with interaction capacity. 
We first train an interaction policy via SL, then we use this learned interaction policy to train the navigation policy via DRL.
As shown in Fig.~\ref{network}, the navigation policy network is composed of two modules, i.e.,
\begin{itemize}
\item planning module, serves as a collision avoidance component in a typical path planning system for a mobile robot, which observes the environment via its surrounding sensor and pedestrian information and outputs expected linear and angular velocities to navigate the robot,
\item interaction module, specifies an interaction policy for robot to decide when to interact with humans, i.e., beep or speak, given its surrounding pedestrian information and its next control commands. 
\end{itemize}

\subsection{Planning Module.}

The input for the planning module is composed of three parts, i.e., the sensor information, the pedestrian information, and the relative goal for the robot. 
In particular, the sensor information is specified by an egocentric local grid map of the robot, which represents the environmental information around the robot, including its shape and observable appearances of obstacles.
The local grid map is constructed from a costmap that is generated by outputs of a 2D laser. 
The pedestrian information consists of three channels of pedestrian maps, which indicate the location and speed of pedestrians around the robot. 
The relative goal specifies the target position and orientation of the robot. 

In this paper, we implement the approach on a differential drive robot that follows desired speed commands. 
Then the output for the planning module consists of a linear velocity $v$ and an angular velocity $\omega$.
We implement both discrete and continuous actions for using multiple DRL methods. 
In specific, for discrete actions, we set a linear velocity $v\in \{0.0,\, 1.0\}$ and an angular velocity $\omega\in \{-0.8,\, -0.4,\, 0.0,\, 0.4,\, 0.8\}$.
For continuous actions, we set $v\in [0,\, 0.6]$ and $\omega\in [-0.9, \, 0.9]$.
Both discrete and continuous actions can be directly performed by the differential robot in our experiments. 
Note that, $v\geq 0$, i.e., moving backward is not allowed, due to the lack of rear sensors. 
As shown in Fig.~\ref{network}, for discrete actions, the network outputs a 10-dimensional vector from a softmax layer to choose the pair of linear and angular velocities.
For continuous actions, the network outputs the mean of linear (resp. angular) velocity sampled from a Gaussian distribution. 

Notice that, the learned interaction policy does not directly change the inputs of the planning module. 
It affects the training of the planning module by affecting behaviors of surrounding pedestrians in the training simulation environments for the planning module in the context of DRL. 
Then we can only ask the navigation policy to minimize the arriving time of the robot without collision. 
When the policy-based DRL algorithm, i.e., PPO~\cite{PPO}, is applied in the planning module, we follow the reward specified in~\cite{yao2021crowd}.

The network for the planning module would be trained in a customized simulator.
In our experiments, both the valued-based DRL algorithm, i.e., V-learning~\cite{CADRL}, and the policy-based DRL algorithm, i.e., PPO, would be implemented and evaluated in the simulator for the discrete and continuous actions, respectively.

\subsection{Interaction Module.}

When to beep or alarm is vital for the interaction policy in social robot navigation, which requires the robot to better understanding the social situation.
A easy policy $P^1_{beep}$ is to perform the beep action $A_{beep}$, when $d_{min}$ (the distance from the robot to the nearest pedestrian) is smaller than a fixed distance ${d_{\theta}}$, i.e.,
\begin{equation}\label{eq:1}
\begin{aligned}
P^1_{beep} &= \begin{cases} A_{beep}, & \text{if } d_{min}<d_\theta ,\\
\textit{Do nothing}, &  \text {otherwise}. 
\end{cases}\\
\end{aligned}
\end{equation}

We can construct an improved policy $P^2_{beep}$ by further considering the speed and orientation of pedestrians, i.e.,
\begin{equation}\label{eq:2}
\begin{aligned}
P^2_{beep} &=  \begin{cases} A_{beep} &\text{if } \forall p.\  ||\mathbf{P}_{p}||<d_\theta \land  ||\mathbf{V}_{p}||>v_\theta \\
                                       & \ \ \ \ \ \land \  \mathbf{P}_{p}\cdot \mathbf{V}_{p}>0 ,\\
\textit{Do nothing}, & \text{otherwise}.
\end{cases}\\
\end{aligned}
\end{equation}
Where $\mathbf{P}_{p}$ is the relative position between the robot and the pedestrian~$p$, $\mathbf{V}_{p}$ denotes the velocity of the pedestrian, and $v_\theta$ is the velocity threshold of the pedestrian.

Above simple interaction policies require predefined fixed thresholds, which may cause unnatural behaviors and are not robust for various dense environments.
Nishimura and Yonetani introduce L2B~\cite{L2B} that designs a delicate reward function to let the robot learn to beep in the context of DRL. 

To better understand the social situation, we propose to learn the interaction policy from the interaction data labeled by human, where the social conventions could be implicitly learned via SL.
Moreover, the interaction policy should not only perform the beep action when it has already got stuck, but also try its best to prevent the robot from freezing. 
In this sense, the next control command of the robot, i.e., the expected linear and angular velocities, needs to be considered as an input of the network for the interaction policy and should be considered when the data labeling as well. 

As shown in Fig.~\ref{network}, after training the network by the labeled interaction data, we can integrate the learned interaction policy to the network for the navigation policy and train the integrated navigation policy in the context of DRL. 
Notice that, in the training process of DRL, the robot is able to observe multiple steps of following states after the beep action.
Then the integrated navigation policy would be able to learn to cooperate with the interaction policy to prevent it from freezing. 
We will evaluate such performances in the experiments.


As shown in Fig.~\ref{network}, the convolutional neural network for binary classification is used in the interaction module for SL.
The network consists of three convolutional layers to extract features hidden in the pedestrian map.
Then, the refined features are merged with the robot's control command from the planning module and passed to a fully-connected layer, which outputs a 2-dimensional vector from a softmax layer to choose whether beep or not.

\subsection{Simulation Environments.}

\begin{figure}
        \centering
        \subfigure[Random scenario]{\includegraphics[width = 0.42\linewidth]{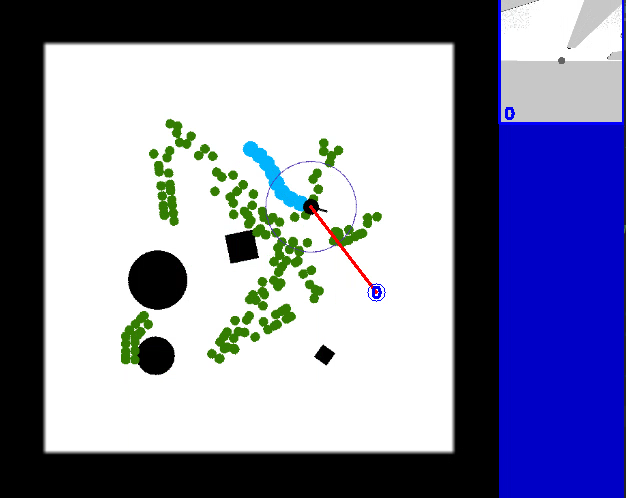}\label{fig:example:e1}}
        \subfigure[Circular scenario]{\includegraphics[width = 0.42\linewidth]{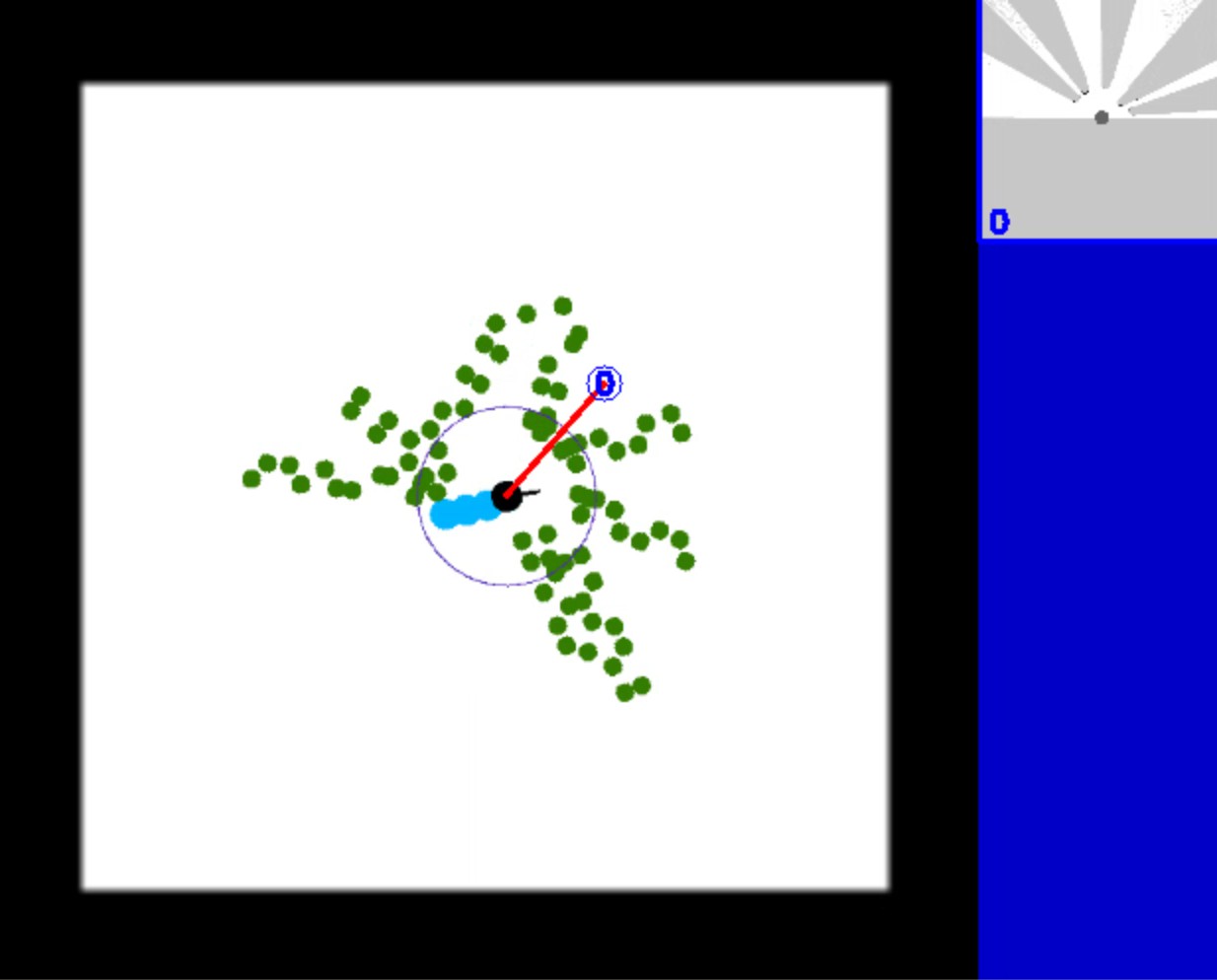}\label{fig:example:e2}}
        \caption{Two scenarios for the training, where the small black circle and light blue spots behind it denote the running robot and its trajectories, 
        green spots denote the legs' trajectories of pedestrians, other black boxes and circles denote static obstacles, and
        the blue circle line around the robot denotes the warning zone for the beep action w.r.t. nearby pedestrians.
        }        
        \label{fig:example:env}
\end{figure}

We train both the planning module and the interaction module in simulation environments generated by a customized 2D simulator based on OpenCV.
Following the discussion in~\cite{yao2021crowd,chen2020distributed}, training the navigation policy in two kinds of scenarios, i.e., random and circular scenarios as shown in Fig.~\ref{fig:example:env}, can lead to a robust policy that not only performs well in various simulation environments but also in the real world. 

In this paper, we train both the planning and interaction policies in these two scenarios.
In particular, a random scenario contains one robot, eight pedestrians, and four static obstacles, where their initial positions and ending positions (for the robot and pedestrians) would be placed randomly.
Meanwhile, a circular scenario contains one robot and eight pedestrians, where all of them would be placed randomly in a circle with a random radius.

Following L2B\footnote{\url{https://github.com/denkiwakame/Python-ERVO}.}, all pedestrians in our simulator are driven by Emotional Reciprocal Velocity Obstacles (ERVO)~\cite{ERVO}, an emotional version of RVO which allows pedestrians to react with the beep sound naturally, where pedestrians' max speed is set to be 1 $m/s$.

\section{Experiments}

In this section, we evaluate our approach for social navigation with interaction capacity in both the simulation and the real world. 
We first specify details of our implementation on both the planning module and the interaction module. 
Then we compare our approach with others, including the ones with simple interaction policies $P^1_{beep}$ and $P^2_{beep}$, and L2B. 
We demonstrate the robustness of the approach by implementing it on different DRL algorithms. 
We also deploy the trained model to a differential drive robot and test its navigation performance in the real world.
Both qualitative and quantitative experiments show that our approach performs well in pedestrian-rich environments.

\subsection{Supervised Learning and Reinforcement Learning Setup.}

We start with the training of the interaction module via SL. 
The labeled training data contains 10,000 pairs of $(x, y)$,
where $x$ is a triple $\langle M_P, v, \omega\rangle$, $M_P$ denotes the pedestrian information which is specified by three channels of the grayscale images to represent the location and speed of pedestrians around the robot, $v$ and $\omega$ denote the linear and angular velocities of the robot, and $y$ is the label for whether beep or not. 
These training data are collected from various simulation environments in the two scenarios, i.e., random scenario and circular scenario. 
We label these data manually.
For DRL, we use PPO as the policy-based method, and V-learning as the value-based method, the parameters of V-learning are all the same as in SARL\footnote{\url{https://github.com/vita-epfl/CrowdNav}.}. 
Parameters for SL and PPO are listed in Table~\ref{table1}.

\begin{table}[t]
        \caption{Parameters for SL and DRL}\label{table1}
        \centering
        {\scriptsize
        \newcommand{\tabincell}[2]{\begin{tabular}[t]{@{}#1@{}}#2\end{tabular}}
        \begin{tabular}[t]{rlrl}
                \toprule
                SL hyper-parameters                                             & Value                                  &
                DRL hyper-parameters &
                Value \\
                \midrule
                \tabincell{c}{dataset size}              & \tabincell{c}{\textbf{$10000$}} &
                \tabincell{c}{learning rate for policy}                &
                 \tabincell{c}{\textbf{$5 \times 10^{-5}$}}
                         \\
                \tabincell{c}{batch size} & \tabincell{c}{\textbf{$1024$}}       &  
                \tabincell{c}{learning rate for value}                 &
                 \tabincell{c}{\textbf{$1 \times 10^{-3}$}}
                \\
                \tabincell{c}{image size}                              & \tabincell{c}{\textbf{$48 \times 48$}}  &
                \tabincell{c}{discount factor ($\gamma$})              & \tabincell{c}{\textbf{$0.99$}} 
                \\
                \tabincell{c}{pedestrian radius}                       & \tabincell{c}{\textbf{$0.3$}}         &
                \tabincell{c}{replay buffer size } & \tabincell{c}{\textbf{$2048$}}
                 \\
                \tabincell{c}{learning rate of Adam}                &
                \tabincell{c}{\textbf{$1 \times 10^{-4}$}} &
                \tabincell{c}{image size}                              & \tabincell{c}{\textbf{$48 \times 48$}} 
                \\
                \tabincell{c}{training epoch}                      & \tabincell{c}{\textbf{$10000$}}          &
                \tabincell{c}{maximum episode length}                      & \tabincell{c}{\textbf{$200$}} 
                \\
                \bottomrule
        \end{tabular}
        }%
\end{table}


In our experiments, both SL and DRL are trained in an i9-9900k CPU and an NVIDIA Titan RTX GPU.

\subsection{Experiments on Simulation Environments.}

\subsubsection{Comparison with simple interaction policies}

We compare the performance of our approach, i.e., SLI, with the ones that replacing the learning interaction policy by simple policies $P^1_{beep}$ and $P^2_{beep}$ respectively, i.e., FD($d_\theta$) and FDV($d_\theta$, $v_\theta$),
where FD($d_\theta$) denotes the social navigation policy that the robot would beep when a pedestrian gets close to a distance less than $d_\theta$ as defined in Eq.~\eqref{eq:1}, 
and FDV($d_\theta$, $v_\theta$) considers both the distance $d_\theta$ and the velocity $v_\theta$. 
Note that, the velocity threshold only considers pedestrians who move towards the robot as specified in Eq.~\eqref{eq:2}.
We set $ d_\theta = 1.0\, m $ as the distance threshold and choose $ v_\theta \in \{ 0.3,\, 0.5,\, 0.7\}\, m/s $ as possible velocity thresholds.
In addition, we use Base to denote the baseline approach that omits the interaction module in SLI.
Here we apply PPO in the planning module for the above methods, which outputs continuous actions of the robot.

Table~\ref{table: fixed threshold} shows the average results of these methods in 500 testing environments of each scenario, where `Success' denotes the ratio of arriving at the goal safely, `PedColl' denotes the ratio of collisions with pedestrians, and `Beep' denotes the ratio of beep times in all steps.

\begin{table}[h]
        \centering
        \caption{Performance of Different Interaction Methods}
        \newcommand{\tabincell}[2]{\begin{tabular}[t]{@{}#1@{}}#2\end{tabular}}
        \renewcommand\arraystretch{0.7}
        \begin{tabular}[t]{clccc}
                \toprule     
                scenarios & Methods   &Success   &PedColl   &Beep \\ 
                
                \midrule
                \multirow{6}*{Random} &
                \tabincell{c}{Base}    & \tabincell{c}{0.668} & \tabincell{c}{0.330}   & \tabincell{c}{-} 
                \\
                \specialrule{0em}{1pt}{1pt}
                & \tabincell{c}{FD(1.0)}    & \tabincell{c}{\textbf{0.956}} & \tabincell{c}{\textbf{0.010}}   & \tabincell{c}{0.624} 
                \\
                \specialrule{0em}{1pt}{1pt}
                & \tabincell{c}{FDV(1.0, 0.3)}    & \tabincell{c}{0.912} & \tabincell{c}{0.056}   & \tabincell{c}{0.120} 
                \\
                \specialrule{0em}{1pt}{1pt}
                & \tabincell{c}{FDV(1.0, 0.5)}    & \tabincell{c}{0.866} & \tabincell{c}{0.100}   & \tabincell{c}{0.077} 
                \\
                \specialrule{0em}{1pt}{1pt}
                & \tabincell{c}{FDV(1.0, 0.7)}    & \tabincell{c}{0.804} & \tabincell{c}{0.138}   & \tabincell{c}{\textbf{0.034}} 
                \\\specialrule{0em}{1pt}{1pt}

                & \tabincell{c}{SLI}    & \tabincell{c}{0.884} & \tabincell{c}{0.062}   & \tabincell{c}{0.097} 
                \\ 
                \specialrule{0em}{1pt}{1pt}

                \midrule

                \multirow{6}*{Circular} &
                \tabincell{c}{Base}    & \tabincell{c}{0.558} & \tabincell{c}{0.442}   & \tabincell{c}{-} 
                \\
                \specialrule{0em}{1pt}{1pt}
                & \tabincell{c}{FD(1.0)}    & \tabincell{c}{0.778} & \tabincell{c}{0.222}   & \tabincell{c}{0.823}  
                \\
                \specialrule{0em}{1pt}{1pt}
                & \tabincell{c}{FDV(1.0, 0.3)}    & \tabincell{c}{0.830} & \tabincell{c}{0.170}   & \tabincell{c}{0.256} 
                \\
                \specialrule{0em}{1pt}{1pt}
                & \tabincell{c}{FDV(1.0, 0.5)}    & \tabincell{c}{0.732} & \tabincell{c}{0.268}   & \tabincell{c}{0.254} 
                \\
                \specialrule{0em}{1pt}{1pt}
                & \tabincell{c}{FDV(1.0, 0.7)}    & \tabincell{c}{0.766} & \tabincell{c}{0.222}   & \tabincell{c}{0.179} 
                \\
                \specialrule{0em}{1pt}{1pt}
                & \tabincell{c}{SLI}    & \tabincell{c}{\textbf{0.864}} & \tabincell{c}{\textbf{0.136}}   & \tabincell{c}{\textbf{0.158}} 

                \\
                \bottomrule 
                \label{table: fixed threshold}
        \end{tabular}
\end{table}

Table~\ref{table: fixed threshold} shows interaction policies can greatly improve the performance of robot navigation in both random and circular scenarios. 
Moreover, our approach, SLI, performs well in environments with high-frequent interactions, like environments in the circular scenario.
Note that, SLI leads to a much lower beep rate than the fixed-threshold methods, i.e., FD and FDV, while keeping a high success rate.
Fig. ~\ref{fig:circle_traj} demonstrates the differences of the beep policy between FDV(1.0, 0.5) and SLI.
When the robot is surrendered by a crowd, FDV beeps frequently by the fixed-threshold policy $P^2_{beep}$, while SLI achieves the similar performance with only few beeps.

For environments in the random scenario, SLI still leads to an acceptable performance compared with others. 
Meanwhile, proper values of $d_\theta$ and $v_\theta$ need to be manually adjusted to achieve better performance. 
These experimental results show that a proper interaction policy can greatly improve the performance of robot navigation in pedestrian-rich environments with very few interactions.

\subsubsection{Comparison with L2B-SARL}

\begin{figure}[htp]
        \centering
        \subfigure[FDV$(1.0, 0.5)$]{\includegraphics[width = 0.44\linewidth]{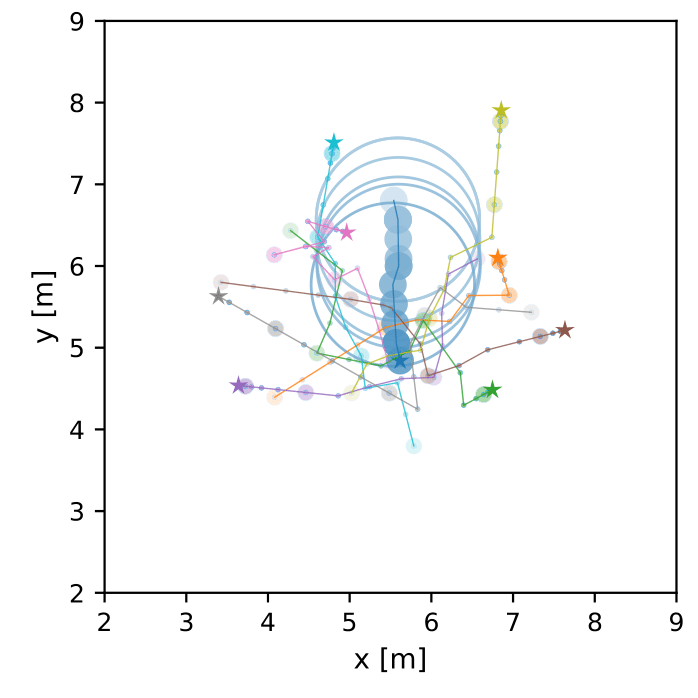}\label{fig:circle_traj:ppo_0.5v}}
        \subfigure[PPO-SLI]{\includegraphics[width = 0.44\linewidth]{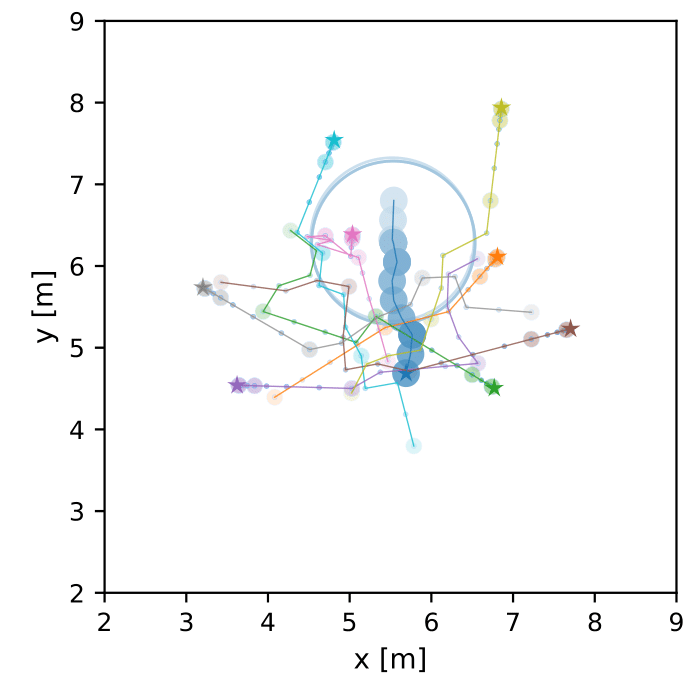}\label{fig:circle_traj:ppo_DRSL}}
        \subfigure[L2B-SARL]{\includegraphics[width = 0.44\linewidth]{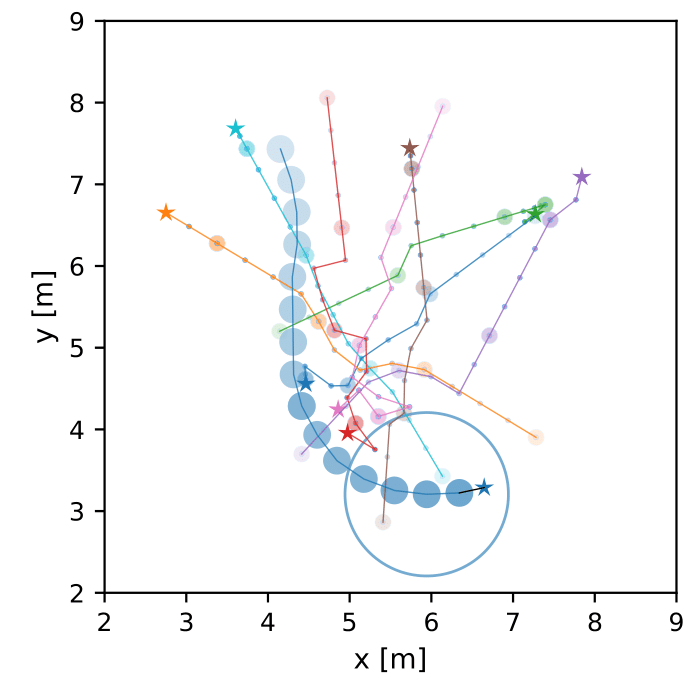}\label{fig:circle_traj:l2b}}
        \subfigure[V-SLI]{\includegraphics[width = 0.44\linewidth]{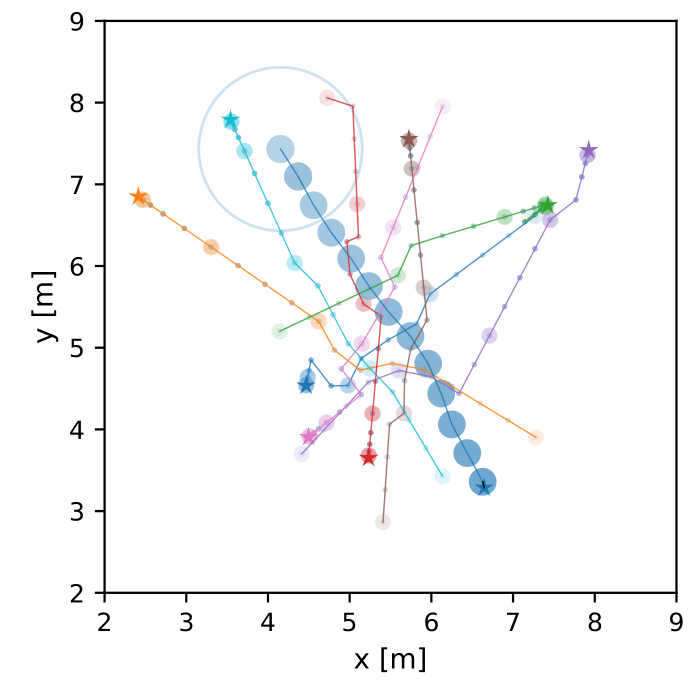}\label{fig:circle_traj:SLI}}
        \caption{The illustration of trajectories generated by various social navigation approaches in the circular scenario with one robot and eight pedestrians. 
        The gradient blue spots denote the trajectory of the running robot, the blue circle around the robot denotes the beep action, and lines with other colors denote different pedestrians, where stars denote their target points.}
        \label{fig:circle_traj}
\end{figure}

Nishimura and Yonetani introduce L2B that designs a delicate reward function to let the robot learn to beep in the context of DRL.
They propose L2B-SARL as a combination of L2B and SARL, i.e., a self-attention mechanism.

We implement L2B-SARL and train it on our simulator. 
For a fair comparison, we also implement V-SLI, which is extended from L2B-SARL by replacing the policy for the beep action with the interaction policy trained by SL in our approach. 
Note that, both approaches output discrete actions of the robot. 

For a fair comparison, we remove static obstacles in environments of the random scenario here, as L2B-SARL does not support static obstacles.
In the experiments, the network parameters were pre-trained via imitation learning, i.e., 6k episodes by ORCA firstly, which due to the fact that the reward function of L2B does not encourage the robot to move towards the goal.

Both quantitative and qualitative results are presented in the following, which show a clear superiority of our method. 

Table~\ref{table:l2b} lists the numerical results of the comparison experiments under different pedestrian numbers $N$ in 500 testing environments of each scenario.
The results show that V-SLI performs better than L2B-SARL in all tests in the sense of success rate and collision rate, while it leads to a slightly higher frequency of the beep action.
With the increase of the number of pedestrians, the performance of L2B-SARL decreases rapidly, while V-SLI still performs well.
Moreover, due to the existence of saddle points in state space about the value function,
L2B-SARL has a higher stuck rate than ours, since their action space is twice as large as ours.

\begin{table}[h]
        \centering
        \caption{Performance of L2B-SARL and V-SLI}
        \newcommand{\tabincell}[2]{\begin{tabular}[t]{@{}#1@{}}#2\end{tabular}}
        \renewcommand\arraystretch{0.5}
        \begin{tabular}[t]{clcccc}
                \toprule     
                scenarios & Methods &N   &Success &PedColl &Beep \\
                \midrule
                \multirow{8}*{Random} &
                \tabincell{c}{L2B-SARL} & \tabincell{c}{4}  & \tabincell{c}{0.824} & \tabincell{c}{0.168}   & \tabincell{c}{\textbf{0.052}}  
                \\
                \specialrule{0em}{1pt}{1pt}
                & \tabincell{c}{V-SLI} & \tabincell{c}{4} & \tabincell{c}{\textbf{0.920}} & \tabincell{c}{\textbf{0.080}}   & \tabincell{c}{0.065}  
                \\
                \specialrule{0em}{1pt}{1pt}
                & \tabincell{c}{L2B-SARL} & \tabincell{c}{6}   & \tabincell{c}{0.686} & \tabincell{c}{0.198}   & \tabincell{c}{\textbf{0.046}} 
                \\
                \specialrule{0em}{1pt}{1pt}
                & \tabincell{c}{V-SLI} & \tabincell{c}{6}  & \tabincell{c}{\textbf{0.882}} & \tabincell{c}{\textbf{0.108}}   & \tabincell{c}{0.098}  
                \\
                \specialrule{0em}{1pt}{1pt}
                & \tabincell{c}{L2B-SARL} & \tabincell{c}{8}   & \tabincell{c}{0.508} & \tabincell{c}{0.198}   & \tabincell{c}{\textbf{0.048}} 
                \\\specialrule{0em}{1pt}{1pt}

                & \tabincell{c}{V-SLI} & \tabincell{c}{8} & \tabincell{c}{\textbf{0.770}} & \tabincell{c}{\textbf{0.144}}   & \tabincell{c}{0.094} 
                \\ 
                \specialrule{0em}{1pt}{1pt}
                
                & \tabincell{c}{L2B-SARL} & \tabincell{c}{10} & \tabincell{c}{0.678} & \tabincell{c}{0.286}   & \tabincell{c}{\textbf{0.051}} 
                \\\specialrule{0em}{1pt}{1pt}

                & \tabincell{c}{V-SLI}  & \tabincell{c}{10}  & \tabincell{c}{\textbf{0.840}} & \tabincell{c}{\textbf{0.100}}   & \tabincell{c}{0.106} 
                \\ 
                \specialrule{0em}{1pt}{1pt}

                \midrule

                \multirow{8}*{Circular} &
                \tabincell{c}{L2B-SARL}  & \tabincell{c}{4} & \tabincell{c}{0.694} & \tabincell{c}{0.296}   & \tabincell{c}{\textbf{0.044}} 
                \\
                \specialrule{0em}{1pt}{1pt}
                & \tabincell{c}{V-SLI} & \tabincell{c}{4} & \tabincell{c}{\textbf{0.808}} & \tabincell{c}{\textbf{0.190}}   & \tabincell{c}{0.081} 
                \\
                \specialrule{0em}{1pt}{1pt}
                & \tabincell{c}{L2B-SARL} & \tabincell{c}{6}   & \tabincell{c}{0.546} & \tabincell{c}{0.442}   & \tabincell{c}{\textbf{0.047}} 
                \\
                \specialrule{0em}{1pt}{1pt}
                & \tabincell{c}{V-SLI} & \tabincell{c}{6} & \tabincell{c}{\textbf{0.656}} & \tabincell{c}{\textbf{0.342}}   & \tabincell{c}{0.109} 
                \\
                \specialrule{0em}{1pt}{1pt}
                & \tabincell{c}{L2B-SARL} & \tabincell{c}{8}  & \tabincell{c}{0.520} & \tabincell{c}{\textbf{0.270}}   & \tabincell{c}{\textbf{0.047}} 
                \\\specialrule{0em}{1pt}{1pt}

                & \tabincell{c}{V-SLI} & \tabincell{c}{8} & \tabincell{c}{\textbf{0.612}} & \tabincell{c}{0.378}   & \tabincell{c}{0.179} 
                \\ 
                \specialrule{0em}{1pt}{1pt}
                
                & \tabincell{c}{L2B-SARL}& \tabincell{c}{10}  & \tabincell{c}{0.514} & \tabincell{c}{\textbf{0.278}}   & \tabincell{c}{\textbf{0.046}} 
                \\\specialrule{0em}{1pt}{1pt}

                & \tabincell{c}{V-SLI}  & \tabincell{c}{10}  & \tabincell{c}{\textbf{0.562}} & \tabincell{c}{0.360}   & \tabincell{c}{0.103} 
                \\ 
                \specialrule{0em}{1pt}{1pt}

                \bottomrule
                \label{table:l2b}

        \end{tabular}
\end{table}

Fig.~\ref{fig:circle_traj} illustrates the difference of interaction policies between L2B-SARL and V-SLI. 
At the beginning of the trajectory,
there is a pedestrian near the robot, L2B-SARL chooses to change the robot's direction to avoid the collision, which leads the robot to take a big turn and meet more pedestrians.
Meanwhile, V-SLI predicts the collision and decides to beep at once, then efficiently moves straight forward to the goal.

\subsubsection{Generalizable experiments}

\begin{figure}[htp]
        \centering
        \begin{overpic}[width=2.7in]{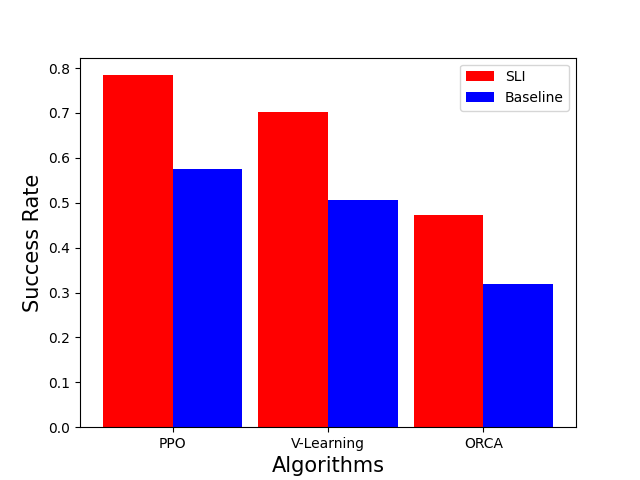}

        \end{overpic}
        \caption{Success rate of PPO-SLI, V-SLI and ORCA-SLI with respect to their baselines.
        The average results are tested in 1000 episodes which consisted of 500 environments of the random scenario and 500 environments of the circle scenario, with 10 pedestrians.
        }
        \label{fig:generalizable}
\end{figure}

We implement the SL interaction module on both the value-based method, i.e., V-learning and the policy-based method, i.e., PPO,
and we also implement the SL interaction module on a conventional model-based method, i.e., ORCA.
The performance of these approaches, i.e., \textit{success rate}, is shown in Fig.~\ref{fig:generalizable}, where the baselines denote the original methods without the SL interaction module.

Fig.~\ref{fig:generalizable} shows that the SL interaction module can be easily adapted to multiple navigation methods and improve their performance. 
The common promotion for baselines with interaction module demonstrates that our approach is generalizable and advantageous.

\subsection{Experiments on the Real Robot.}

\begin{figure}[htp]
        \centering
        \begin{overpic}[width=0.95\linewidth]{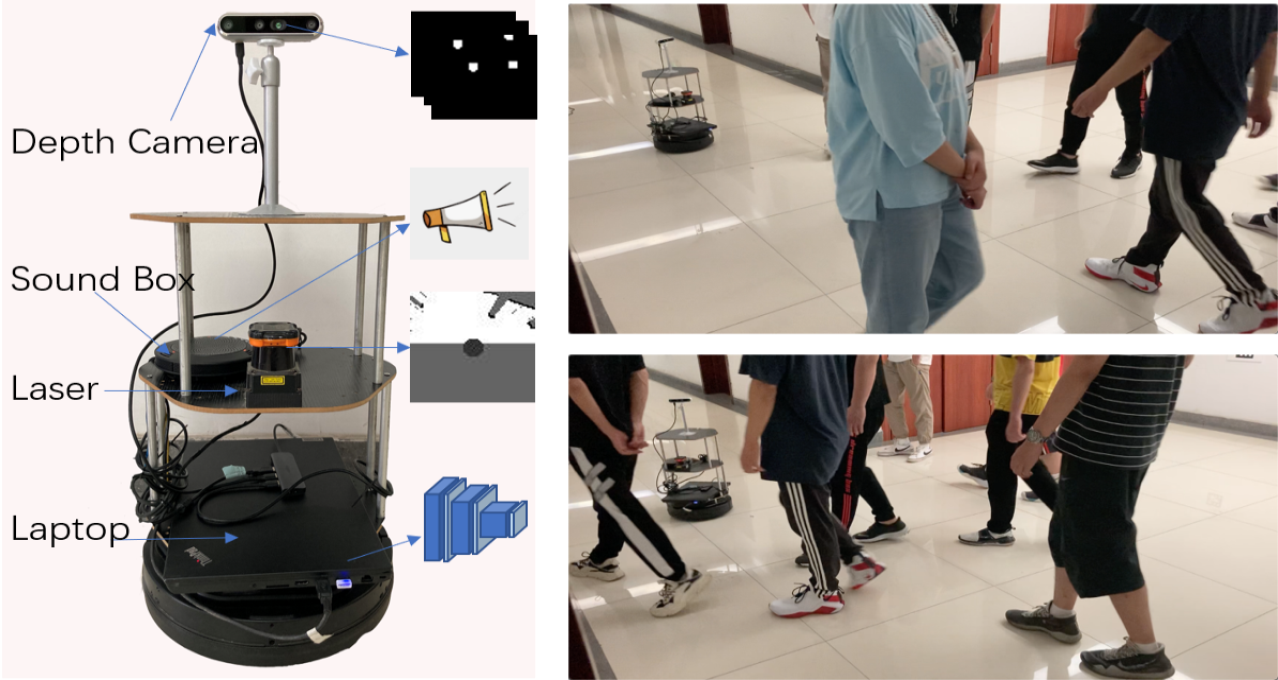}

        \end{overpic}
        \caption{The robot in our experiments, where the 3-channel pedestrian map is generated from the depth camera,
        the sensor map is from the laser sensor, the laptop is for inferring the deep neural network (left).
        A pedestrian-rich environment (right) and the beeping robot (bottom-right) are illustrated.
        }
        \label{fig:real_robot}
\end{figure}

\begin{figure}[htp]
        \centering
        \begin{overpic}[width = 0.95\linewidth]{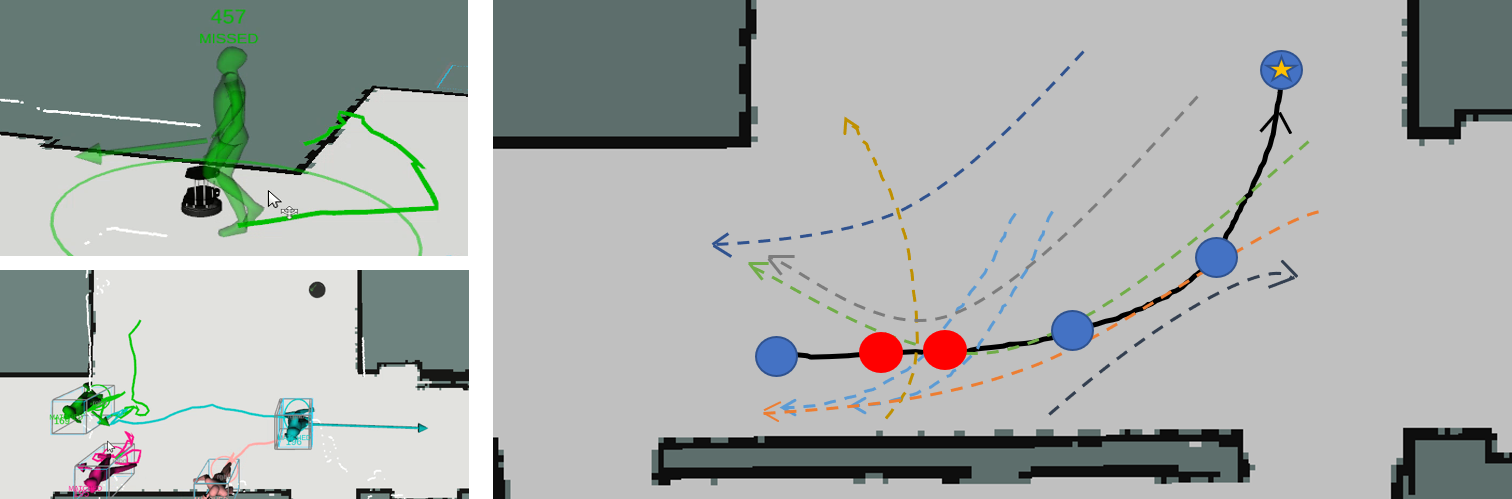}
        \end{overpic}
        \caption{The illustration of the real-world robot, which tracks persons through combining yolo and spencer people tracking (left). The trajectory of the robot (blue cycle) in a crowded environment (right) is illustrated, where
        the red circle denotes it's beep action.}
        \label{fig:real_traj}
\end{figure}

As shown in Fig.~\ref{fig:real_robot}, we use Kobuki base TurtleBot 2 as the real-world robot, which is equipped with a Hokuyo UTM-30LX Scanning Laser Rangefinder as the 2D laser sensor, 
a RealSense D455 depth camera as the pedestrian detection and tracking sensor, as well as a Meeteasy Mvoice 1000 as the audio output.
In addition, we choose ThinkPad P15v as the computation device (Intel core i7-11800H CPU, 8cores, 16threads, 4.6GHZ, Nvidia-T600 4GB GPU, 2.1kg),
which allows the real robot to run the control model at 63 HZ. 
We deploy the trained policy by PPO-SLI to the robot since it outperforms others including V-SLI.
We use yolo-V3~\cite{redmon2018yolov3} to detect pedestrians and spencer people tracking~\cite{linder2016multi} to track pedestrians, as shown in Fig.~\ref{fig:real_traj}.

We test our method in a crowded corridor environment, as shown in Fig.~\ref{fig:real_robot}, the robot is carrying out a navigation task while pedestrians walking in the vicinity.
Note that, some humans may ignore the robot, i.e., is using the mobile phone, which increases the difficulty of navigation task. 
Fig.~\ref{fig:real_traj} illustrates the trajectory of the robot in the environment, which demonstrates that our approach allows the robot reach the goal efficiently and beep in a proper way while people are about to surround it.
Furthermore, when obscured by pedestrians, proper interactions to remind pedestrians to move away are beneficial for the robot to restore its localization. 


\section{Conclusion}
\label{conclusion}

We propose a learning approach for robot social navigation with interaction capacity in pedestrian-rich environments, where the interaction policy, SLI, is first trained via SL, then it incorporates with a DRL based navigation method to generate the social navigation policy.
We evaluate the approach on both simulation environments and a real-world robot.
Both quantitative and qualitative experimental results show that, the SLI policy can be easily adapted to multiple navigation methods and improve their performance, which allows the robot to navigate through crowds efficiently.

%


\bibliographystyle{IEEEtran}

\end{document}